\documentclass{article}

\PassOptionsToPackage{numbers}{natbib}
\usepackage[preprint]{nips_2018}

\usepackage[utf8]{inputenc} % allow utf-8 input
\usepackage[T1]{fontenc}    % use 8-bit T1 fonts
\usepackage{hyperref}       % hyperlinks
\usepackage{url}            % simple URL typesetting
\usepackage{booktabs}       % professional-quality tables
\usepackage{amsfonts}       % blackboard math symbols
\usepackage{amsmath}
\usepackage{nicefrac}       % compact symbols for 1/2, etc.
\usepackage{microtype}      % microtypography
\usepackage{graphicx}
\usepackage{color}
\usepackage{rotating}
\usepackage{multirow}
\usepackage{varwidth}

\title{\texttt{geomstats}: a Python Package for Riemannian Geometry in Machine Learning}

\author{
Nina Miolane \\
  Stanford University\\
  Stanford, CA 94305 \\
  \texttt{nmiolane@stanford.edu} \\
  \And
  Johan Mathe \\
  Froglabs AI\\
  San Francisco, CA94103, USA\\
  \texttt{johan@froglabs.ai} \\
  \And
   Claire	Donnat \\
    Stanford University			 \\
    Stanford, CA 94305, USA \\
  \texttt{cdonnat@stanford.edu}
    \And
    Mikael	Jorda	\\
    Stanford University			 \\
    Stanford, CA 94305, USA \\
    \texttt{mjorda@stanford.edu} \\
    \And
    Xavier Pennec \\
    Inria Sophia-Antipolis \\
    06902 Valbonne, France \\
    \texttt{xavier.pennec@inria.fr}
}

\begin{document}

\maketitle

\begin{abstract}
We introduce \texttt{geomstats}, a python package that performs computations on manifolds such as hyperspheres, hyperbolic spaces, spaces of symmetric positive definite matrices and Lie groups of transformations. We provide efficient and extensively unit-tested implementations of these manifolds, together with useful Riemannian metrics and associated Exponential and Logarithm maps. The corresponding geodesic distances provide a range of intuitive choices of Machine Learning's loss functions. We also give the corresponding Riemannian gradients. The operations implemented in \texttt{geomstats} are available with different computing backends such as \texttt{numpy}, \texttt{tensorflow} and  \texttt{keras}. We have enabled GPU implementation and integrated \texttt{geomstats}' manifold computations into  \texttt{keras}' deep learning framework. This paper also presents a review of manifolds in machine learning and an overview of the \texttt{geomstats} package with examples demonstrating its use for efficient and user-friendly Riemannian geometry.
%\textbf{Python code:} \texttt{https://github.com/anonymous\_authors}.
\end{abstract}

%%% XP: not sure to understand excatly the meaning of "unit-tested implementations of these manifolds"

%\tableofcontents

\section{Introduction}

There is a growing interest in using Riemannian geometry in machine learning. To illustrate the reason for this interest, consider a standard supervised learning problem: given an input $X$, we want to predict an output $Y$. We can model the relation between $X$ and $Y$ by a function $f_\theta$ parameterized by a parameter $\theta$. There are three main situations where Riemannian geometry can naturally appear in this setting: through the input $X$, the output $Y$, or the parameter $\theta$. For example the input $X$ can belong to a Riemannian manifold or be an image defined on a Riemannian manifold. The input $X$ can also be a manifold itself, for example a 2D smooth surface representing a shape such as a human pose \cite{Bronstein2017}. Similarly, the output $Y$ may naturally belong to a Riemannian manifold, as in \cite{Hou2018} where a neural network is used to predict the pose of a camera which is an element of the Lie group $SE(3)$. Finally, the parameter $\theta$ of a model can be constrained on a Riemannian manifold as in the work of \cite{Huang2017} which constrains the weights of a neural network on multiple dependent Stiefel manifolds.

%In these three situations, the manifolds are known in advance and are used to model or constrain $X$, $Y$ and $\theta$.

%If the manifolds are not known in advance, they can be learned through an unsupervised learning step. In unsupervised learning, the goal is to uncover the hidden structure of a set of inputs $X$. In the context of manifold learning, we assume that the inputs belong to an underlying lower dimensional manifold. We aim to uncover this manifold through methods such as multidimensional scaling (MDS) \cite{Tenenbaum2000}, locally linear embedding (LLE) \cite{Roweis2000}, or stochastic neighbor embedding (t-SNE) \cite{vanDerMaaten2008}, among others. Then, we get a manifold on which to constrain inputs $X$ for example for future supervised learning algorithms.

There are intuitive and practical advantages for modeling inputs, outputs and parameters on manifolds. Computing on a lower dimensional space leads to manipulating fewer degrees of freedom, which can potentially imply faster computations and less memory allocation. Moreover, the non-linear degrees of freedom that arise in a lower dimensional space often make more intuitive sense: cities on the earth are better localized giving their longitude and latitude, i.e., their manifold coordinates, than giving their position $x, y, z$ in the 3D space.

Yet, the adoption of Riemannian geometry by the larger machine learning community has been inhibited by the lack of a modular framework for implementing such methods. Code sequences are often custom tailored for specific problems, and are not easily reused. To address this issue, some packages have been written to perform computations on manifolds. The \texttt{theanogeometry} package \cite{theanogeometry} provides an implementation of differential geometric tensors on manifolds where closed forms do not necessarily exist, using the automatic differentiation tool \texttt{theano} to integrate differential equations that define the geometric tensors. The \texttt{pygeometry} package \cite{pygeometry} offers an implementation primarily focused on the Lie groups $SO(3)$ and $SE(3)$ for robotics applications. However, there is no implementation of non-canonical metrics on these Lie groups. The \texttt{pymanopt} package \cite{pymanopt} (which builds upon the matlab package \texttt{manopt} \cite{Boumal2014} but is otherwise independent) provides a very comprehensive toolbox for optimization on a extensive list of manifolds. Still, the choice of metrics is restricted on these manifolds which are often implemented using canonical embeddings in higher-dimensional euclidean spaces.

% TODO: There is a need ... / There is a lack of .... .

This paper presents \texttt{geomstats}, a package specifically targeted to the machine learning community to perform computations on Riemannian manifolds with a flexible choice of Riemannian metrics. The \texttt{geomstats} package makes four contributions. First, \texttt{geomstats} is the first Riemannian geometry package to be extensively unit-tested with more than 90 \% code coverage. Second, \texttt{geomstats} implements \texttt{numpy} \cite{numpy} and \texttt{tensorflow} \cite{tensorflow} backends, making computations intuitive, vectorized for batch computations, and available for GPU implementation. Third, we provide an updated version of the \texttt{keras} deep learning framework equipped with Riemannian gradient descent on manifolds. Fourth, \texttt{geomstats} has an educational role on Riemannian geometry for computer scientists that can be used as a \textit{complement} to theoretical papers or books. We refer to \cite{Postnikov2001} for the theory and expect the reader to have a high-level understanding of Riemannian geometry.

An overview of \texttt{geomstats} is given in Section~\ref{sec:geomstats}. We then present concrete use cases of \texttt{geomstats} for machine learning on manifolds of increasing geometric complexity, starting with manifolds embedded in flat spaces in Section~\ref{sec:hypersphere}, to a manifold embedded in a Lie group with a Lie group action in Section~\ref{sec:spdspace}, to the Lie groups $SO(n)$ and $SE(n)$ in Section~\ref{sec:se3}. Along the way, we present a review of the occurrences of each manifold in the machine learning literature, some educational visualizations of the Riemannian geometry as well as implementations of machine learning models where the inputs, the outputs and the parameters successively belong to manifolds.

\section{The Geomstats Package}\label{sec:geomstats}

\subsection{Geometry}

The \texttt{geomstats} package implements Riemannian geometry using a natural object-oriented approach with two main families of classes: the manifolds, inherited from the class \texttt{Manifold} and the Riemannian metrics, inherited from the class \texttt{RiemannianMetric}. Children classes of \texttt{Manifold} considered here include: \texttt{LieGroup}, \texttt{EmbeddedManifold}, \texttt{SpecialOrthogonalGroup}, \texttt{SpecialEuclideanGroup}, \texttt{Hypersphere}, \texttt{HyperbolicSpace} and \texttt{SPDMatricesSpace}. Then, the Riemannian metrics can equip the manifolds. Instantiations of the \texttt{RiemannianMetric} class and its children classes are attributes of the manifold objects.

The class \texttt{RiemannianMetric} implements the usual methods of Riemannian geometry, such as the inner product of two tangent vectors at a base point, the (squared) norm of a tangent vector at a base point, the (squared) distance between two points, the Riemannian Exponential and Logarithm maps at a base point and a geodesic characterized by an initial tangent vector at an initial point or by an initial point and an end point. Children classes of \texttt{RiemannianMetric} include the class \texttt{InvariantMetric}, which implements the left- and right- invariant metrics on Lie groups.

The methods of the above classes have been extensively unit-tested, with more than 90\% code coverage. The code is provided with \texttt{numpy} and \texttt{tensorflow} backends. The code is vectorized through the use of arrays, to facilitate intuitive batch computations. The \texttt{tensorflow} backend also enables running the computations on GPUs.

\subsection{Statistics and Machine Learning}

The package \texttt{geomstats} also implements the ``Statistics" aspect of Geometric Statistics - specifically Riemannian statistics through the class \texttt{RiemannianMetric} \cite{Pennec2006}. The class \texttt{RiemannianMetric} implements the weighted Fr\'echet mean of a dataset through a Gauss-Newton gradient descent iteration \cite{Frechet1948}, the variance of a dataset with respect to a point on the manifold, as well as tangent principal component analysis \cite{Fletcher2004}.

The package facilitates the use of Riemannian geometry in machine learning and deep learning settings. Suppose we want to train a neural network to predict on a manifold, \texttt{geomstats} provides off-the-shelf loss functions on Riemannian manifolds, implemented as squared geodesic distances between the predicted output and the ground truth. These loss functions are consistent with the geometric structure of the Riemannian manifold. The package gives the closed forms of the Riemannian gradients corresponding to these losses, so that back-propagation can be easily performed.

Suppose we want to constrain the parameters of a model, for example the weights of a neural network, to belong to a manifold. We provide modified versions of \texttt{keras} and \texttt{tensorflow}, so that they can constrain weights on manifolds during training.

% TODO(johmathe): Say a bit more here.

In the following sections, we demonstrate the use of the manifolds implemented in \texttt{geomstats}. For each manifold, we present a literature review of its appearance in machine learning and we describe its implementation in \texttt{geomstats} together with a concrete use case.

\section{Embedded Manifolds - Hypersphere and Hyperbolic Space}\label{sec:hypersphere}

We consider the hypersphere and the hyperbolic space, respectively implemented in the classes \texttt{Hypersphere} and \texttt{HyperbolicSpace}. The logic of the Riemannian structure of these two manifolds is very similar. They are both manifolds defined by their embedding in a flat Riemannian or pseudo-Riemannian manifold.

The $n$-dimensional hypersphere $S^{n}$ is defined by its embedding in the $(n+1)$-Euclidean space, which is a flat Riemannian manifold, as
\begin{equation}
S^{n} = \left\{
x \in \mathbb{R}^{n+1}: x_1^2 + ... + x_{n+1}^2 = 1
\right\}.
\end{equation}

Similarly, the $n$-dimensional hyperbolic space $H_n$ is defined by its embedding the $(n+1)$-dimensional Minkowski space, which is a flat pseudo-Riemannian manifold, as
\begin{equation}
H_{n} = \left\{
x \in \mathbb{R}^{n+1}: - x_1^2 + ... + x_{n+1}^2 = -1
\right\}.
\end{equation}

The classes \texttt{Hypersphere} and \texttt{HyperbolicSpace} therefore inherit from the class \texttt{EmbeddedManifold}. They implement methods such as: conversion functions from intrinsic $n$-dimensional coordinates to extrinsic $(n+1)$-dimensional coordinates in the embedding space (and vice-versa); projection of a point in the embedding space to the embedded manifold; projection of a vector in the embedding space to a tangent space at the embedded manifold.

The Riemannian metric defined on $S^n$ is derived from the Euclidean metric in the embedding space, the Riemannian metric defined on $H^n$ is derived from the Minkowski metric in the embedding space. They are respectively implemented in the classes \texttt{HypersphereMetric} and \texttt{HyperbolicMetric}.

\subsection{Hypersphere - Use Cases Review in Machine Learning}

We review the contexts where it is natural to embed data on a hypersphere. Examples include circular statistics, directional statistics or orientation statistics which focus on data on circles, spheres and rotation groups. Applications are obviously extremely diverse and include biology and physics, among many others \cite{Mardia2000}. In biology, the sphere $S^2$ is used in analysis of protein structures \cite{Kent2005}. In physics, the semi-hypersphere $S^4_+$ is used to encode the projective space $P_4$ for representing crystal orientations in applied crystallography \cite{Schaeben1993}.

The shape statistics literature is also manipulating data on abstract hyperspheres. Kendall's studies shapes of $k$ landmarks in $m$ dimensions and introduces the pre-shape spaces which are hyperspheres $S^{m(k-1)}$ \cite{Kendall1989}. The s-rep, a skeletal representation of 3D shapes, also deals with hyperspheres $S^{3n-4}$ as the object under study is represented by $n$ points along its boundary \cite{Hong2016}.

Lastly, hyperspheres can be used to constrain the parameters of a machine learning model. For example, training a neural net with parameters constrained on a hypersphere results in an easier optimization, faster convergence and comparable (even better) classification accuracy \cite{Liu2017}.

\subsection{Geomstats Use Case - Optimization and Deep Learning on Hyperspheres}\label{sec:sn}

We demonstrate the use of \texttt{geomstats} for constraining neural networks' weights on manifolds during training following the deep learning literature \cite{Liu2017}. The folder \texttt{deep\_learning} of the supplementary materials contains the implementation of this use case.

\begin{figure}[h!]
\vspace{-4mm}
    \centering
    \includegraphics[width=200pt]{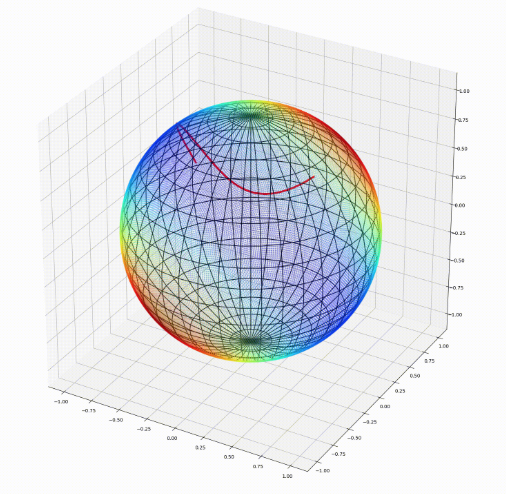}
    \caption{Minimization of a scalar field on the sphere $S^2$. The color map indicates the scalar field values, where blue is the minimum and red the maximum. The red curve shows the trajectory taken by the Riemannian gradient descent, which converges to a minimum (blue region).}
    \label{fig:hypersphere}
\end{figure}

First, however, we provide the implementation of the Riemannian gradient descent on the hypersphere. Our example minimizes a quadratic form $x^TAx$ with $A\in  \mathbb{R}_{n \times n}$ and $x^TAx > 0$ constrained on the hypersphere $S^{n-1}$. Geomstats allows us to conveniently generate a positive semidefinite matrix by doing a random uniform sampling on the \texttt{SPDMatricesSpace} manifold. Figure~\ref{fig:hypersphere} illustrates the Riemannian optimization process.

As for neural network's training, the optimization step has been modified in \texttt{keras} such than the stochastic gradient descent is done on the manifold through the Exponential map. In our implementation, the user can pass a \texttt{manifold} parameter to each neural network layer. The stochastic gradient descent optimizer has been modified to operate the Riemannian gradient descent in parallel. It infers the number of manifolds directly from the dimensionality by finding out how many manifolds are needed in order to optimize the number of kernel weights of a given layer.

We provide a modified version of a simple deep convolutional neural network and a resnet \cite{resnet2015} with its convolutional layers' weights trained on the hypersphere. They were trained respectively on the MNIST \cite{mnist} and \cite{cifar10} datasets.

\subsection{Hyperbolic Space - Use Case Reviews in Machine Learning}

We review the machine learning literature that deals with Hyperbolic spaces. Hyperbolic spaces arise in information and learning theory. The space of univariate Gaussian endowed with the Fisher metric densities is a hyperbolic space. This characterization is used in various fields, and for example in image processing where each image pixel is represented by a Gaussian distribution \cite{Angulo2014} and in radar signal processing where the corresponding echo is represented by a stationary Gaussian process \cite{Arnaudon2013}.

The hyperbolic spaces can also be seen as continuous versions of trees and are therefore interesting when learning hierarchical representations of data \cite{Nickel2017}. Hyperbolic geometric graphs (HGG) have also been suggested as a promising model for social networks, where the hyperbolicity appears through a competition between similarity and popularity of an individual \cite{Papadopoulos2012}.

\subsection{Geomstats Use Case - Visualization on the Hyperbolic space $H_2$}\label{sec:hn}

\begin{figure}[h!]
    \centering
    \includegraphics[width=190pt]{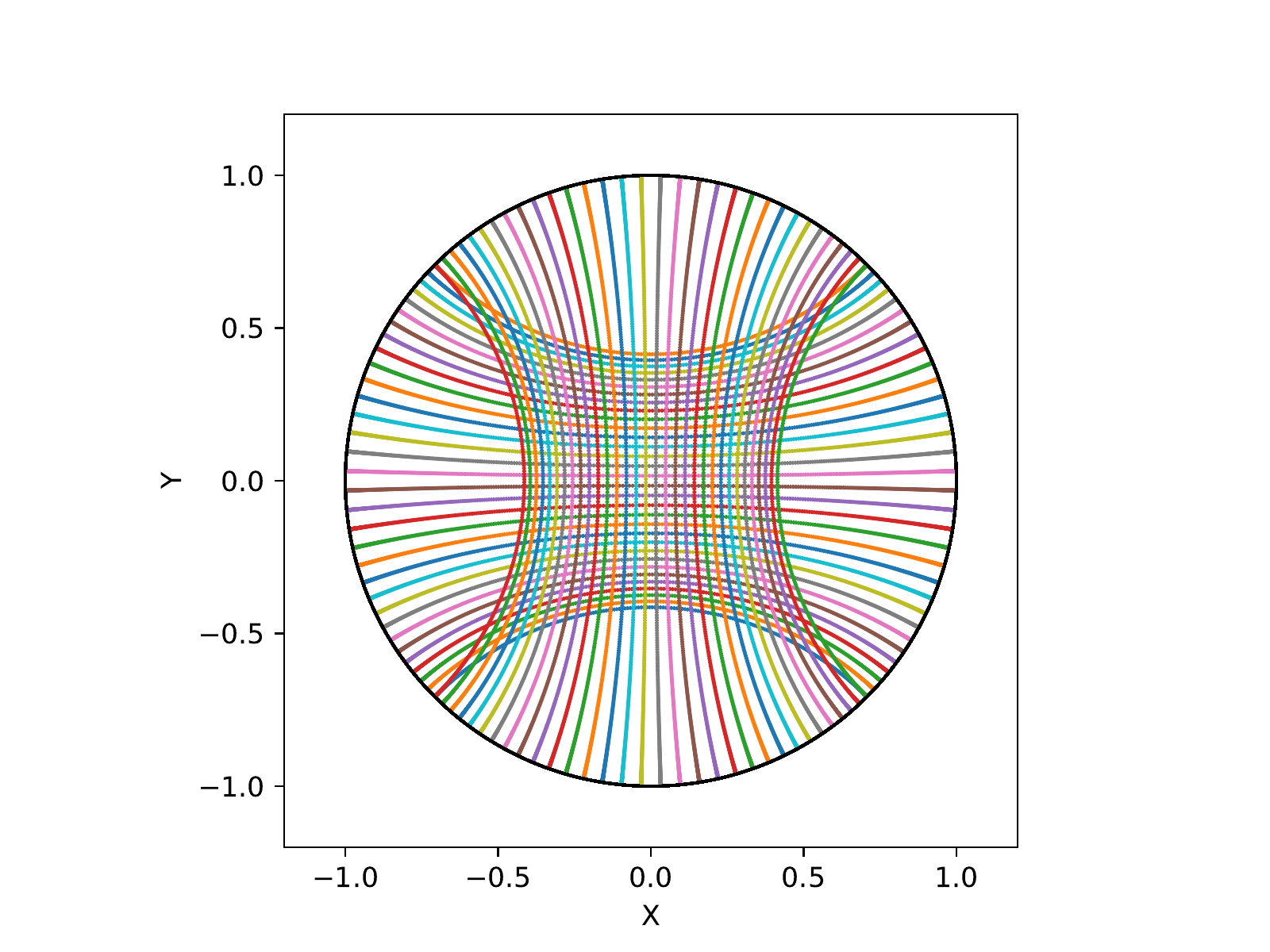}
    \includegraphics[width=190pt]{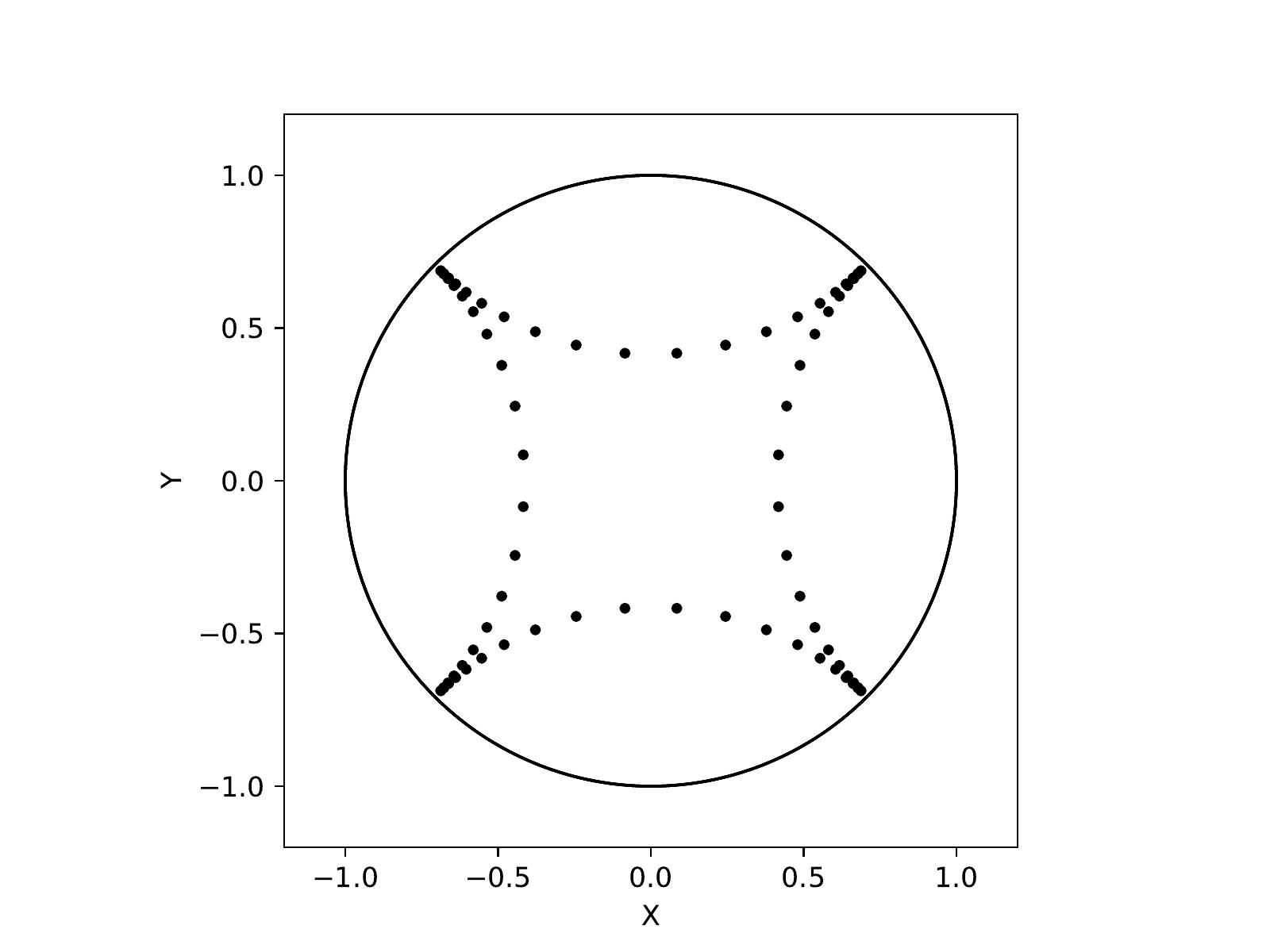}
    \caption{Left: Regular geodesic grid on the Hyperbolic space $H^2$ in Poincare disk representation. Right: Geodesic square on the Hyperbolic space $H_2$, with points regularly spaced on the geodesics defining the square's edges.}
    \label{fig:poincaredisk}
\end{figure}

We present the visualization toolbox of \texttt{geomstats}, that plays an educational role by enabling the users to test their intuition on Riemannian manifolds. They can run and adapt the examples provided in the \texttt{geomstats/examples} folder of the supplementary materials. For example, we can visualize the hyperbolic space $H_2$ through the Poincare disk representation, where the border of the disk is at infinity. The user can then observe how a geodesic grid and a geodesic square are deformed in the hyperbolic geometry on Figure~\ref{fig:poincaredisk}.

\section{Manifold of Symmetric Positive Definite (SPD) Matrices}\label{sec:spdspace}

We have seen the Hypersphere and the Hyperbolic space that are manifolds embedded in flat spaces. Now we increase the geometric complexity and consider a manifold embedded in the General Linear group of invertible matrices. The manifold of symmetric positive definite (SPD) matrices in $n$ dimensions is indeed defined as
\begin{equation}
SPD = \left\{
S \in \mathbb{R}_{n \times n}: S^T = S, \forall z \in \mathbb{R}^n, z \neq 0, z^TSz > 0
\right\}.
\end{equation}
The class \texttt{SPDMatricesSpace} thus inherits from the class \texttt{EmbeddedManifold} and has an \texttt{embedding\_manifold} attribute which stores an object of the class \texttt{GeneralLinearGroup}. We equip the manifold of SPD matrices with an object of the class \texttt{SPDMetric} that implements the affine-invariant Riemannian metric of \cite{Pennec2006b} and inherits from the class \texttt{RiemannianMetric}.

\subsection{SPD Matrices Manifold - Use Cases in Machine Learning}

SPD matrices are used for data representation in many fields \cite{Cherian2016}. This subsection lists their use cases. In diffusion tensor imaging (DTI), the ``diffusion tensors" are ellipsoids that are 3x3 SPD matrices at each voxel. They spatially characterize the diffusion of water molecules in the tissues. These fields of SPD matrices are inputs to regression models, for example of an intrinsic local polynomial regression applied to comparison of fiber tracts of HIV subjects compared with a control group in \cite{Yuan2012}.

In functional magnetic resonance imaging (fMRI), the brain connectome framework extracts connectivity graphs from a set of patients' resting-state images' time series \cite{sporns2005human,wang2013disrupted,ingalhalikar2014sex}. The regularized graph Laplacians of the respective graphs form a dataset of SPD matrices. They represent a compact summary of the brain's connectivity patterns which is used to assess neurological responses to a variety of stimuli (drug, pathology, patient's activity, etc.).

In medical imaging and computational anatomy, SPD matrices can also encode anatomical shape changes observed in images. The SPD matrix $J^{T}J^{1/2}$ represents the directional information of shape change captured by the Jacobian matrix $J$ at a given voxel \cite{Grenander2007}.

Covariance matrices are also SPD matrices which appear in many settings. We find covariance clustering used for sound compression in acoustic models of automatic speech recognition (ASR) systems \cite{Shinohara2010} and covariance clustering for material classification \cite{Faraki2015}, among others. Covariance descriptors are also popular image descriptors or video descriptors \cite{Harandi2014}.

Lastly, SPD matrices have found applications in deep learning, where they are used as features extracted by a neural network. The authors of \cite{Gao2017} show that an aggregation of learned deep convolutional features into a SPD matrix creates a robust representation of images that enables to outperform state-of-the-art methods on visual classification.

\subsection{Geomstats Use Case - Connectivity Graph Classification}\label{sec:spd}

We show through a concrete brain connectome application how  \texttt{geomstats} can be easily leveraged for efficient supervised learning on the space of SPD matrices. The folder \texttt{brain\_connectome} of the supplementary materials contains the implementation of this use case.

We consider the fMRI data from the 2014 MLSP Schizophrenia Classification challenge\footnote{Data openly available at \url{https://www.kaggle.com/c/mlsp-2014-mri}}, consisting of the resting-state fMRIs of 86 patients split into two balanced categories: control vs people suffering schizophrenia. Consistently with the connectome literature, we approach the classification task by using a SVM classifier on the pre-computed pairwise-similarities between brains. The critical step lies in our ability to correctly identify similar brain structures, here represented by regularized Laplacian SPD matrices $\hat{L}=(D-A)+\gamma I$, where A and D are respectively the adjacency and the degree matrices of a given connectome. The parameter $\gamma$ is a regularization shown to have little effect on the classification performance \cite{Dodero2015}.

Following two popular approaches in the literature \cite{Dodero2015}, we define similarities between connectomes through kernels relying on the Riemannian distance $d_R(\hat{L}_1,\hat{L}_2)= ||\log(\hat{L}_1^{-1/2}.\hat{L}_2.\hat{L}_1^{-1/2})||_F$ and on the log-Euclidean distance, a computationally-lighter proxy for the first:
$d_{LED}(\hat{L}_1,\hat{L}_2)= ||\log_{I}(\hat{L}_2) -\log_{I}(\hat{L}_1)||_F$. In these formulae, $\log$ is the matrix logarithm and $F$ refers to the Frobenius norm. Both of these similarities are easily computed with \texttt{geomstats}, for example the Riemannian distance is obtained through \texttt{metric.squared\_dist} where \texttt{metric} is an instance of the class \texttt{SPDMetric}.

\begin{figure}
\vspace*{-5mm}
  \begin{minipage}[c]{0.45\linewidth}
    \centering
    \vspace*{-18mm}
    \begin{tabular}{|c|c|c|} \hline
       Distance & Accuracy & F1-Score \\ \hline \hline
       Riemannian & 30.8\% &\textbf{47.1}\\ \hline
       Log Euclidean & \textbf{62.5}  &36.4 \\ \hline
       Frobenius   &  46. 2\%   & 0.00 \\ \hline
    \end{tabular}
  \end{minipage}
\hfill
  \begin{minipage}[b]{0.45\linewidth}
    \includegraphics[width=6cm]{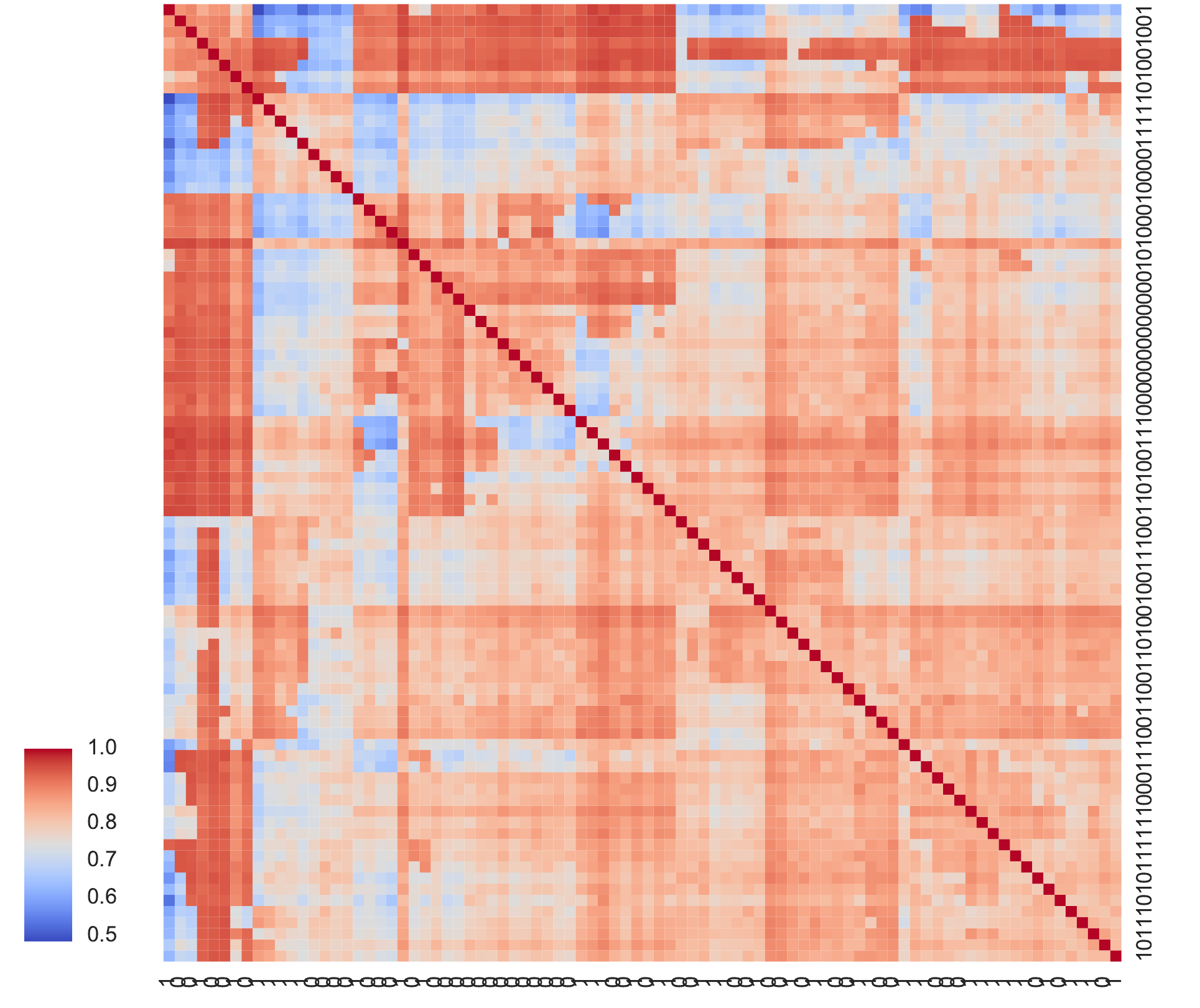}
      \end{minipage}
    \caption{Left: Connectome classification results. Right: Clustermap of the recovered similarities using the Riemannian distance on the SPD Manifold. We note in particular the identification of several clusters (red blocks on the diagonal)}
    \label{fig:SPD}
    \end{figure}

Figure~\ref{fig:SPD} (left) shows the performance of these similarities for graph classification, which we benchmark against a standard Frobenius distance. With an out-of-sample accuracy of 61.2\%, the log-Euclidean distance here achieves the best performance. Interestingly, the affine-invariant Riemannian distance on SPD matrices is the distance that picks up the most differences between connectomes. While both the Frobenius and the log-Euclidean recover only very slight differences between connectomes --placing them almost uniformly afar from each other--, the Riemannian distance exhibits greater variability, as shown by the clustermap in Figure~\ref{fig:SPD} (right). Given the ease of implementation of these similarities with \texttt{geomstats}, comparing them further opens research directions for in-depth connectome analysis.

\section{Lie Groups $SO(n)$ and $SE(n)$ - Rotations and Rigid Transformations}\label{sec:se3}

We have seen manifolds embedded in other manifolds, where the embedding manifolds were either flat or had a Lie group structure. Now we turn to manifolds that are Lie groups themselves. The special orthogonal group $SO(n)$ is the group of rotations in $n$ dimensions defined as
\begin{equation}
SO(n) = \left\{
R \in \mathbb{R}_{n \times n }: R^T.R = Id_n \text{ and } \det R = 1
\right\}.
\end{equation}

The special Euclidean group $SE(n)$ is the group of rotations and translations in $n$ dimensions defined by its homegeneous representation as
\begin{equation}
SE(n) = \left\{ X \in \mathbb{R}_{n \times n } \quad | \quad X = \left[ \begin{array}{c|c} R & t \\ \hline 0 & 1 \end{array} \right], t \in \mathbb{R}^n, R \in SO(n) \right\}
\end{equation}

The classes \texttt{SpecialOrthogonalGroup} and \texttt{SpecialEuclideanGroup} both inherit from the classes \texttt{LieGroup} and \texttt{EmbeddedManifold}, as embedded in the General Linear group. They both have an attribute \texttt{metrics} which can store a list of metric objects, instantiations of the class \texttt{InvariantMetric}. A left- or right- invariant metric object is instantiated through an inner-product matrix at the tangent space at the identity of the group.

\subsection{Lie Groups $SO(n)$ and $SE(n)$ - Use Cases in Machine Learning}

This subsection enumerates the use cases of the Lie groups $SO(n)$ and $SE(n)$ for data and parameters representation. In 3D, $SO(3)$ and $SE(3)$ appear naturally when dealing with articulated objects. A spherical robot arm is an example of articulated object, whose positions can be modeled as the elements of $SO(3)$. The human spine can also be modeled as an articulated object where each vertebra is represented as an orthonormal frame that encodes the rigid body transformation from the previous vertebra \cite{Arsigny:PHD:2006, Boisvert2008}.

In computer vision, elements of $SO(3)$ or $SE(3)$ are used to represent the orientation or pose of cameras \cite{Kendall2015}. Supervised learning algorithm predicting such orientations or poses have numerous applications for robots and autonomous vehicles which need to localize themselves in their environment.

Lastly, the Lie group $SO(n)$ and its extension to the Stiefel manifold, are found very useful in the training of deep neural networks. The authors of \cite{Huang2017} suggest to constrain the network's weights on a Stiefel manifold, i.e. forcing the weights to be orthogonal to each other. Enforcing the geometry significantly improves performances, reducing for example the test error of wide residual network on CIFAR-100 from 20.04\% to 18.61\% .

\subsection{Geomstats Use Case - Geodesics on $SO(3)$}\label{sec:son}

%\textbf{TODO: see if we can get nice properties by changing the metric}

Riemannian geometry can be easily integrated for machine learning applications in robotics applications using \texttt{geomstats}. We demonstrate this by presenting the interpolation of a robot arm trajectory by geodesics. The folder \texttt{robotics} of the supplementary materials contains the implementation of this use case.

In robotics, it is common to control a manipulator in Cartesian space rather than configuration space. This allows for a much more intuitive task specification, and makes the computations easier by solving several low dimension problems instead of a high dimension one. Most robotic tasks require to generate and follow a position trajectory as well as an orientation trajectory.

\begin{figure}[h]
    \centering
    \includegraphics[width = 0.25\textwidth]{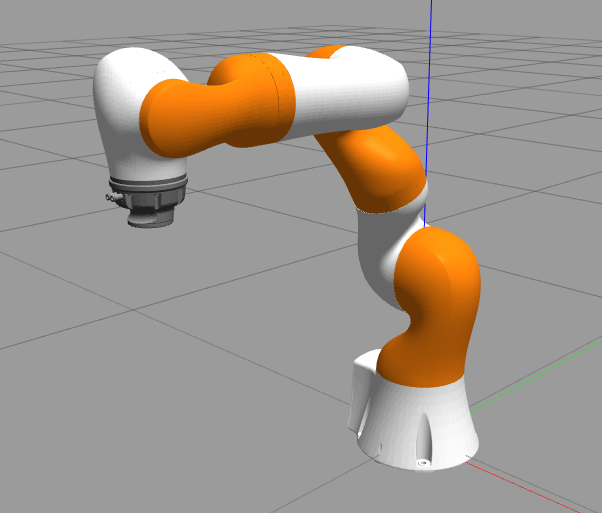}
    \includegraphics[width = 0.25\textwidth]{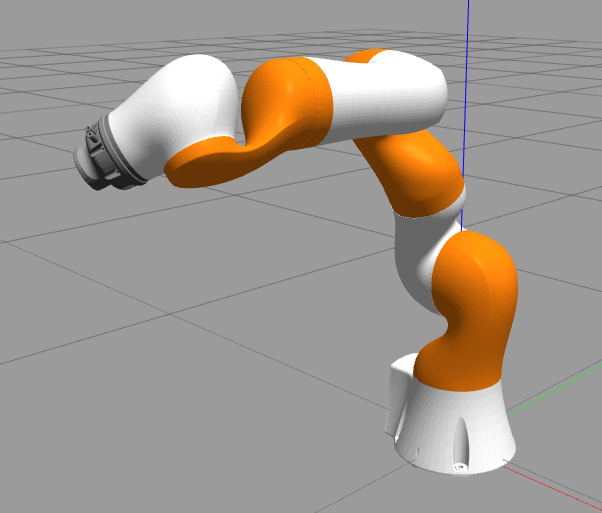}
    \includegraphics[width = 0.25\textwidth]{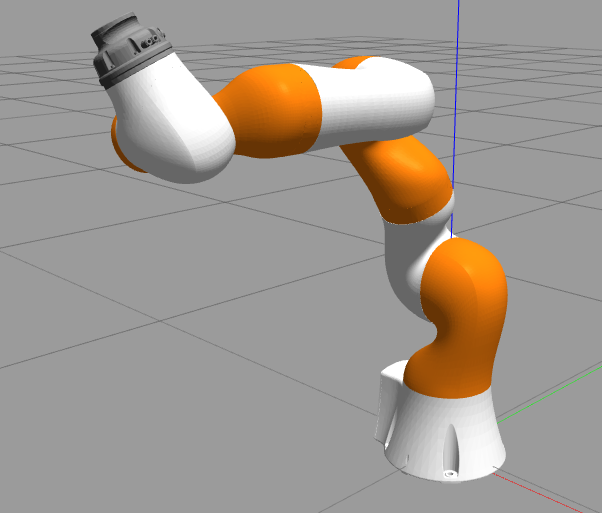}
    \caption{A Riemannian geodesic computed with the canonical bi-invariant metric of $SO(3)$, applied to the extremity of the robotic arm.}
    \label{fig:robot}
\end{figure}

%TODO : put RGB spatial frame at end of robot

While it is quite easy to generate a trajectory for position using interpolation between several via points, it is less trivial to generate one for orientations that are commonly represented as rotation matrices or quaternions. Here, we show that we can actually easily generate an orientation trajectory as a geodesic between two elements of $SO(3)$ (or as a sequence of geodesics between several via points in $SO(3)$). We generate a geodesic on $SO(3)$ between the initial orientation of the robot and its desired final orientation, and use the generated trajectory as an input to the robot controller. The trajectory obtained is illustrated in Figure~\ref{fig:robot}.

This opens the door for research at the intersection of Riemannian geometry, robotics and machine learning. We could ask the robot arm to perform a trajectory towards an element of $SE(3)$ or $SO(3)$ predicted by a supervised learning algorithm trained for a specific task. The next subsection presents the concrete use case of training a neural network to predict on Lie groups using \texttt{geomstats}.

\subsection{Geomstats Use Case - Deep Learning Predictions on $SE(3)$}\label{sec:pose}

We show how to use \texttt{geomstats} to train supervised learning algorithms to predict on manifolds, specifically here: to predict on the Lie group $SE(3)$. This use case is presented in more details in the paper \cite{Hou2018} and the open-source implementation is given. The authors of \cite{Hou2018} consider the problem of pose estimation that consists in predicting the position and orientation of the camera that has taken a picture given as inputs.

The outputs of the algorithm belong to the Lie group $SE(3)$. The \texttt{geomstats} package is used to train the CNN to predict on $SE(3)$ equipped with a left-invariant Riemannian metric. At each training step, they use the loss given by the squared Riemannian geodesic distance between the predicted pose and the ground truth. The Riemannian gradients required for back-propagation are given by the closed forms implemented in \texttt{geomstats}.

\begin{figure}[ht]
\centering
\includegraphics[width=0.75\linewidth]{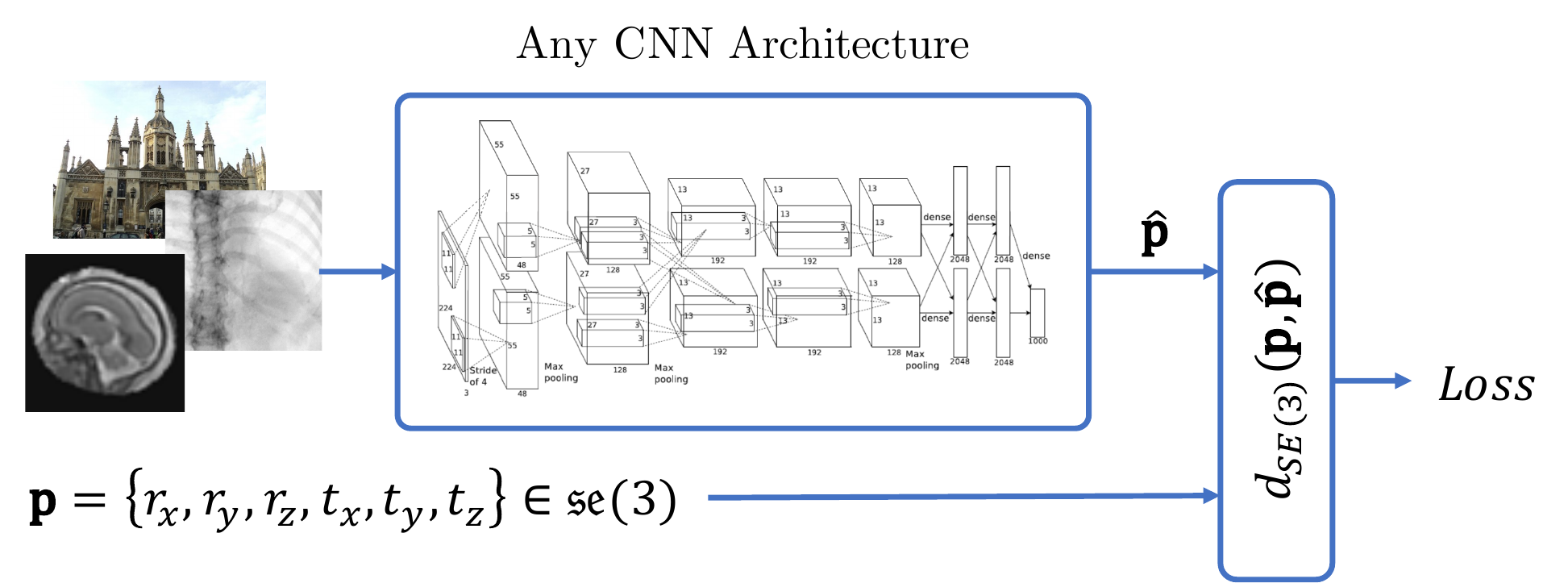}
\caption{Image courtesy of \cite{Hou2018}. CNN with a squared Riemannian distance as the loss on $SE(3)$.}
\label{fig:overview}
\end{figure}

The effectiveness of the Riemannian loss is demonstrated by experiments showing significative improvements in accuracy for image-based 2D to 3D registration. The loss functions and gradients provided in \texttt{geomstats} extend this research directions to CNN predicting on other Lie groups and manifolds.

\section{Conclusion and Outlook}

We introduce the open-source package \texttt{geomstats} to democratize the use of Riemannian geometry in machine learning for a wide range of applications. Regarding the geometry, we have presented manifolds of increasing complexity: manifolds embedded in flat Riemannian spaces, then the case of the SPD matrices space and lastly Lie groups with invariant Riemannian metrics. This provides an educational tool for users who want to delve into Riemannian geometry through a hands-on approach, with intuitive visualizations for example in subsections~\ref{sec:hn} and \ref{sec:son}.

In regard to machine learning, we have presented concrete use cases where inputs, outputs and parameters belong to manifolds, in the respective examples of subsection~\ref{sec:spd}, subsection~\ref{sec:pose} and subsection~\ref{sec:sn}. They demonstrate the usability of \texttt{geomstats} package for efficient and user-friendly Riemannian geometry. Regarding the machine learning applications, we have reviewed the occurrences of each manifold in the literature across many different fields. We kept the range of applications very wide to show the many new research avenues that open at the cross-roads of Riemannian geometry and machine learning.

\texttt{geomstats} implements manifolds where closed-forms for the Exponential and the Logarithm maps of the Riemannian metrics exist. Future work will involve implementing manifolds where these closed forms do not necessarily exist. We will also provide the \texttt{pytorch} backend.

\small

\bibliographystyle{splncs03}
\bibliography{main}

\end{document}